\documentclass[10pt,twocolumn,letterpaper]{article}

\usepackage{cvpr}              

\usepackage{times}  
\usepackage{helvet}  
\usepackage{courier}  
\usepackage[hyphens]{url}  
\usepackage{graphicx} 
\usepackage{natbib}  
\usepackage{caption} 
\usepackage{arydshln} 
\usepackage{booktabs}
\usepackage{graphicx}
\usepackage{amssymb,amsfonts,amsmath}
\usepackage{epsfig}
\usepackage{algorithm}
\usepackage{algorithmic}
\usepackage{booktabs}
\usepackage{mathrsfs}
\usepackage{color}
\usepackage{textcomp}
\usepackage{bbding}
\usepackage{pifont}
\usepackage{wasysym}
\usepackage{subcaption}
\usepackage{xcolor,colortbl}
\usepackage{overpic}
\usepackage{microtype}
\usepackage{multirow}
\usepackage{makecell}
\usepackage{hhline}
\usepackage[american]{babel}
\usepackage{cuted}
\usepackage{ragged2e}
\definecolor{lightgray}{gray}{.94}
\definecolor{tinygray}{gray}{.96}

\usepackage{xcolor}
\usepackage{multirow}
\usepackage{epsfig}
\usepackage{adjustbox}
\usepackage[pagebackref=true,
            breaklinks=true,colorlinks,
            citecolor=blue,linkcolor=red,
            bookmarks=false]{}

\definecolor{cvprblue}{rgb}{0.21,0.49,0.74}
\usepackage[pagebackref,breaklinks,colorlinks,allcolors=cvprblue]{hyperref}

\def\paperID{4267} 
\def\confName{CVPR}
\def\confYear{2026}

\title{Measurement-Constrained Sampling for Text-Prompted Blind Face Restoration}

\author{Wenjie Li$^{1}$, Yulun Zhang$^{2}$, Guangwei Gao$^{3}$, Heng Guo$^{1}$, and Zhanyu Ma$^{1}$\\
$^1$Beijing University of Posts and Telecommunications\\
$^2$Shanghai Jiao Tong University \ \ 
$^3$Nanjing University of Science and Technology\\
{\tt\small \{cswjli, guoheng, mahzhanyu\}@bupt.edu.cn}, 
{\tt\small \{yulun100, csggao\}@gmail.com}
}

\begin{document}
\maketitle

\begin{abstract}
Blind face restoration (BFR) may correspond to multiple plausible high-quality (HQ) reconstructions under extremely low-quality (LQ) inputs. However, existing methods typically produce deterministic results, struggling to capture this one-to-many nature. In this paper, we propose a Measurement-Constrained Sampling (MCS) approach that enables diverse LQ face reconstructions conditioned on different textual prompts. Specifically, we formulate BFR as a measurement-constrained generative task by constructing an inverse problem through controlled degradations of coarse restorations, which allows posterior-guided sampling within text-to-image diffusion. Measurement constraints include both Forward Measurement, which ensures results align with input structures, and Reverse Measurement, which produces projection spaces, ensuring that the solution can align with various prompts. Experiments show that our MCS can generate prompt-aligned results and outperforms existing BFR methods. Codes will be released after acceptance.
\end{abstract}


\section{Introduction}
\begin{figure}[t]
\begin{overpic}[width=0.99\linewidth]{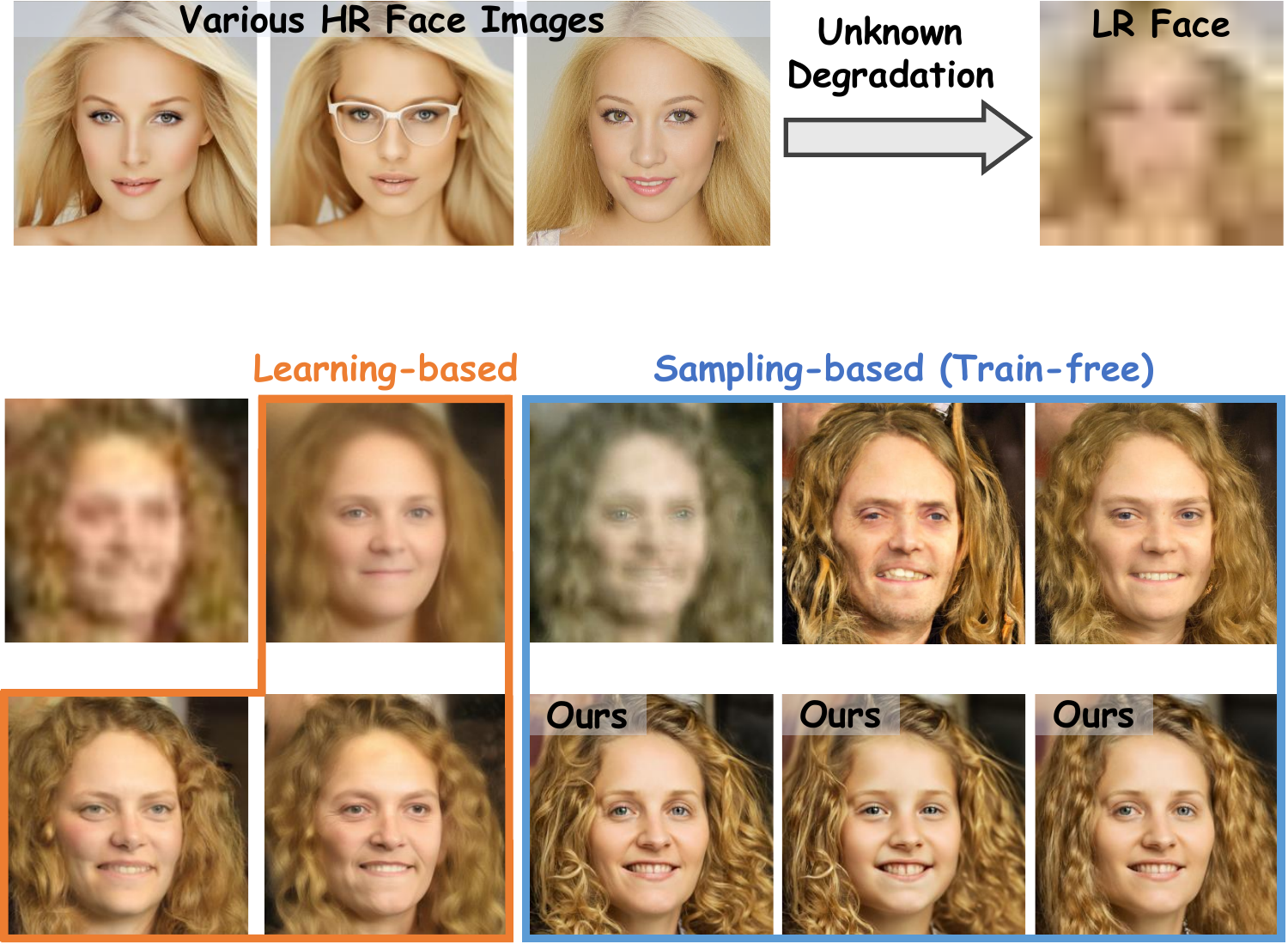}
\put(4.0,20.5){\color{black}{\fontsize{7.5pt}{1pt}\selectfont Real LQ}}
\put(24.0,20.5){\color{black}{\fontsize{7.5pt}{1pt}\selectfont DR2~\cite{wang2023dr2}}}
\put(42.6,20.5){\color{black}{\fontsize{7.5pt}{1pt}\selectfont DDNM~\cite{wang2022zero}}}
\put(62.9,20.5){\color{black}{\fontsize{7.5pt}{1pt}\selectfont PGDiff~\cite{yang2023pgdiff}}}
\put(82.5,20.5){\color{black}{\fontsize{7.5pt}{1pt}\selectfont SSDiff~\cite{li2025self}}}

\put(1.2,-3.2){\color{black}{\fontsize{7.5pt}{1pt}\selectfont DifFace~\cite{yue2024difface}}}
\put(21.9,-3.2){\color{black}{\fontsize{7.5pt}{1pt}\selectfont DiffBIR~\cite{lin2024diffbir}}}
\put(46.5,-3.2){\color{black}{\fontsize{7.5pt}{1pt}\selectfont `Null'}}
\put(63.1,-3.2){\color{black}{\fontsize{7.5pt}{1pt}\selectfont `Little girl'}}
\put(81.4,-3.2){\color{black}{\fontsize{7.5pt}{1pt}\selectfont `Straight hair'}}

\put(5,49.2){\color{black}{\fontsize{9pt}{1pt}\selectfont (a) Different HQ faces may degrade into the same LQ face.}}
\put(0.8,-9.5){\color{black}{\fontsize{9pt}{1pt}\selectfont (b) Our method can generate high-quality prompt-aligned faces.}}

\end{overpic}
\vspace{8mm}
   \caption{Illustration of the one-to-many nature of blind degradations and the effectiveness of our method in generating diverse, prompt-aligned face images from extremely low-quality inputs.}
\label{fig: teaser}
\vspace{-4mm}
\end{figure}

Blind face restoration (BFR) aims to reconstruct high-quality (HQ) facial images from low-quality (LQ) inputs with unknown and complex degradations, which has been extensively studied in recent years~\cite{li2023survey}. This task becomes particularly ambiguous under extremely LQ conditions, where a single input may correspond to multiple semantically plausible HR reconstructions~\cite{li2023survey}. As illustrated in Fig.~\ref{fig: teaser}, different face images with varying attributes (\eg, expressions, hairstyles, accessories) may degrade into visually similar LQ observations. Therefore, this one-to-many nature not only increases the challenge of restoration but also reveals the potential for personalized and controllable BFR.

However, existing BFR~\cite{wang2021towards,wang2023dr2,yang2023pgdiff,lin2024diffbir} methods struggle to capture such one-to-many uncertainty in blind degradations, limiting their flexibility in applications. Specifically, with the increasing adoption of diffusion models~\cite{ho2020denoising} in BFR, surpassing the performance of GAN-based methods~\cite{wang2021towards,zhou2022towards}, existing diffusion-based methods can be broadly classified into two categories: \textbf{\emph{i) Learning-based methods}}~\cite{wang2023dr2,lin2024diffbir} typically rely on supervised training with paired low- and high-quality data, and are often optimized to produce a single deterministic output, making them inadequate for modeling the inherent one-to-many correspondence under severe degradations. \textbf{\emph{ii) Sampling-based methods}}~\cite{wang2022zero,chung2022diffusion,yang2023pgdiff,li2025self} typically accomplish restoration by modifying the sampling trajectory of pre-trained diffusion models, enabling zero-shot, training-free adaptation to a variety of restoration tasks. 


\begin{figure*}[t]
\begin{overpic}[width=1\linewidth]{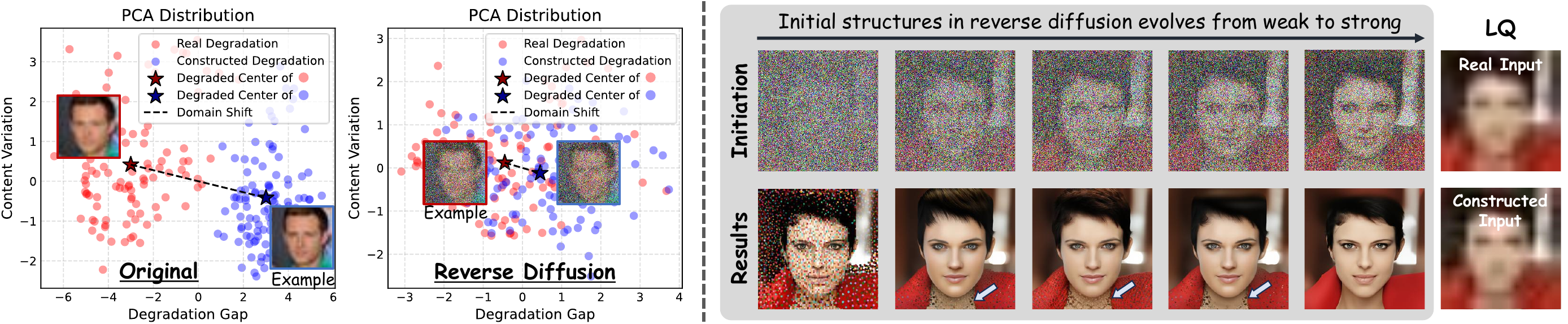}

\end{overpic}
\vspace{-5mm}
   \caption{(\textbf{Left}) While real and constructed degradations differ, the gap diminishes in reverse diffusion, making constructed observations suitable anchors of inverse problem solving for guided sampling. (\textbf{Right}) When facial structures in noisy samples are unclear in the reverse diffusion process of T2I diffusion, inverse problem solving centred on constructed degradations amplifies noise, resulting in artifacts.}
\label{fig: insight}
\vspace{-3mm}
\end{figure*}

Sampling-based approaches utilizing inverse problem solving~\cite{wang2022zero,chung2022diffusion,cardoso2023monte}, though not inherently text-driven, show potential for generating diverse outputs when combined with pre-trained text-to-image (T2I) diffusion models. Specifically, these methods define a degradation operation $A$, guiding each step ${{\hat{\boldsymbol{x}}}_t}$ of the reverse diffusion to degenerate to LQ inputs ${\boldsymbol{I}_{LQ}}$, \eg, $A\left( {{{\hat{\boldsymbol{x}}}_t}} \right) \!\!\to\!\! {\boldsymbol{I}_{LQ}}$. Since HQ faces with varying attributes may map to the same LQ face, this approach, when integrated with text guidance, can produce a solution space with various HQ faces exhibiting diverse attributes. However, its applicability is limited, as it often assumes prior knowledge of the degradation operation $A$, which is challenging to satisfy in blind or real-world degradation contexts.

Building on the successes and limitations mentioned above, we aim to construct a degradation observation that approximates real degradation in the T2I diffusion while enabling the solution of inverse problems. Therefore, we explore the relationship between real degradation and constructed degradation (\eg, bicubic interpolation applied to results after a restoration process input). As shown in the left panel of Fig.~\ref{fig: insight}, during the reverse diffusion process, where noise dominates, the gap between constructed and real degradation significantly narrows. This observation led us to formulate an inverse problem in which constructed degradation can serve as an anchor for reverse diffusion. However, as shown in the right panel of Fig.~\ref{fig: insight}, this formulation can disrupt the reverse diffusion process when facial structures in noisy samples are unclear, resulting in pronounced artifacts. The solution of the inverse problem can be mitigated when facial structures in the noisy samples are well-defined.

Based on these observations, we propose a Measurement-Constrained Sampling (MCS) that unifies text-driven control and measurement guidance within pre-trained T2I diffusion. To solve the inverse problem, we first apply a constructed degradation to the rough estimate from the restorer~\cite{chen2020learning}, creating an anchor. Simultaneously, we introduce a Reverse Measurement to constrain the denoising trajectory within a degradation-consistent projection space, anchored by the constructed degradation and guided by prompts. To avoid artifacts arising from the reverse strategy when facial structures are unclear in denoised samples, as discussed in Sec.~\ref{subsec:Method}, we introduce a Forward Measurement in early sampling steps to approximate restoration that aligns with the structures of inputs. Additionally, we employ noise-blended anchors to initialize reverse diffusion, providing an approximate facial structure and bridging the domain gap between synthesis and real degradation under noise dominance. 

To the best of our knowledge, our MCS is the first BFR method capable of generating text-prompt-aligned faces in a training-free manner. As shown in Fig.~\ref{fig: teaser}, benefiting from balancing Forward and Reverse Measurement, our MCS outperforms existing learning-based~\cite{wang2023dr2,lin2024diffbir,yue2024difface} and sampling-based~\cite{wang2022zero,yang2023pgdiff,li2025self} diffusion methods in both BFR diversity and quality. In summary, our contributions are as follows:

\begin{itemize}
\item We leverage pre-trained T2I models to reconstruct diverse HR faces from blind LR inputs, guided by both text-based semantics and measurement consistency constraints;
\item We introduce forward measurements to approximate degradations like facial structures, and reverse measurements to expand generative spaces for satisfying diversity prompts;
\item Under blind degradation, our MCS produces accurate face results without prompts and generates text-aligned faces when guided, outperforming existing BFR methods.
\end{itemize}

\section{Related Work}
\subsection{Blind Face Restoration}
To avoid the limitations of non-blind methods~\cite{chen2018fsrnet,xin2020facial,li2024efficient,wan2025attention}, which are difficult to generalize to realistic degradations, blind FSR has seen rapid progress in recent years. Early approaches~\cite{li2020blind,wang2022restoreformer} enhanced networks by incorporating face dictionary priors, but these methods lack robustness under extremely LR conditions. To address this, recent works incorporate generative priors into networks. Specifically, GLEAN~\cite{chan2021glean}, GPEN~\cite{yang2021gan}, GFPGAN~\cite{wang2021towards}, and SGPN~\cite{zhu2022blind} integrate StyleGAN-based priors into network architectures to boost performance in specific scenarios. VQFR~\cite{gu2022vqfr}, CodeFormer~\cite{zhou2022towards}, and DAEFR~\cite{tsai2023dual} further enhance the determinism of facial feature representations by introducing discretized GAN priors, such as pre-trained codebooks, thereby reducing the likelihood of facial artifacts. With the advent of generative diffusion~\cite{rombach2022high} in generative tasks, methods like StableSR~\cite{wang2024exploiting}, DiffBIR~\cite{lin2024diffbir}, OSEDiff~\cite{wu2024one}, and OSDFace~\cite{wang2025osdface} improve FSR by training lightweight controllable models to harness the generative diffusion prior. However, these methods require large-scale paired data training and can produce only a single deterministic restoration result. 

\begin{figure*}[t]
\begin{overpic}[width=0.99\linewidth]{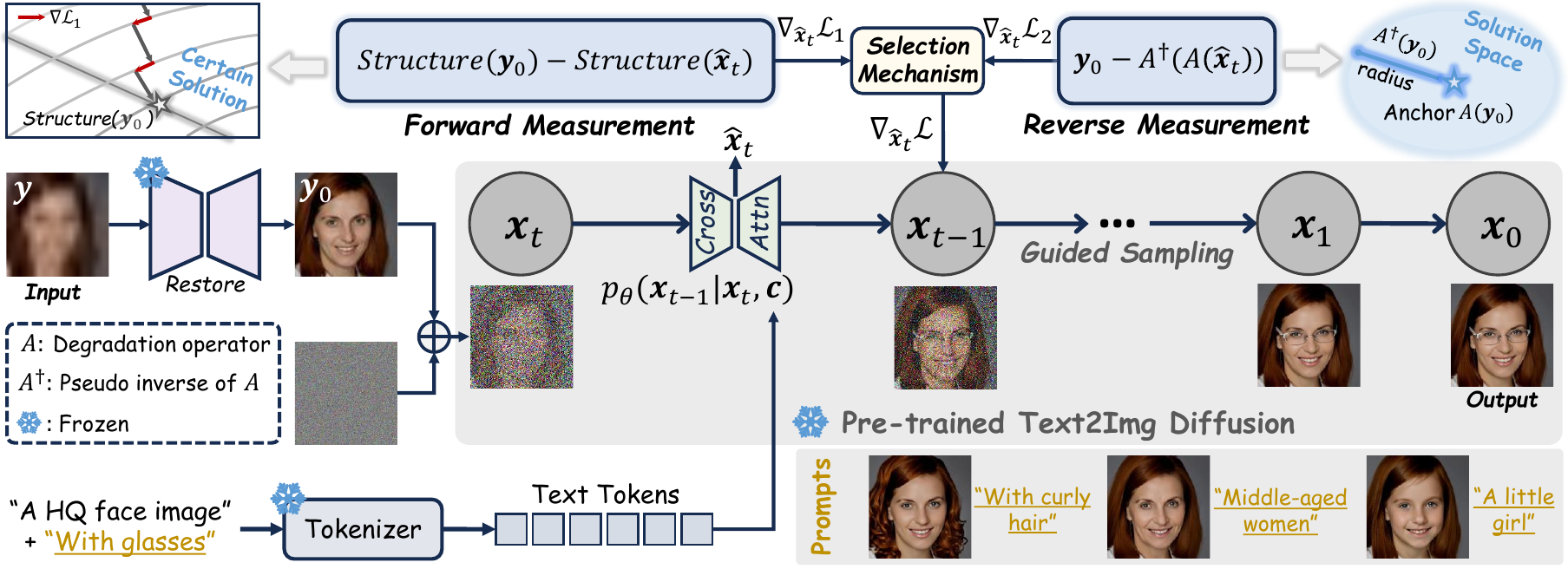}
\end{overpic}
\vspace{0mm}
   \caption{Our method utilizes \emph{Forward Measurement} to align solutions with input structures, where structural components after the wavelet decomposition define the structure. \emph{Reverse Measurement} ensures solutions lie within the degenerate space centered on anchor points. Therefore, under the guidance of texts, our method can generate diverse facial results that satisfy both degeneracy and textual constraints.}
\label{fig: pipeline}
\vspace{-1mm}
\end{figure*}

\subsection{Guided Sampling-based Diffusion} 
Recent efforts have focused on developing efficient sampling-guided strategies for pre-trained diffusion models to enable zero-shot image restoration without training. Early methods such as DDRM~\cite{kawar2022denoising}, IIGDM~\cite{song2023pseudoinverse}, DDNM~\cite{wang2022zero}, FPS~\cite{dou2024diffusion}, and MCGDiff~\cite{cardoso2023monte} rely on known linear degradation models (\eg, bicubic downsampling) and their corresponding inverses to guide the restoration process. Later approaches, including DPS~\cite{chung2022diffusion} and GDP~\cite{fei2023generative}, replace explicit likelihood modeling with hard data consistency constraints to tackle more complex nonlinear degradation, enabling approximate posterior inference during denoising. Gandikota \etal~\cite{gandikota2024text} further explore text-guided restoration using these sampling frameworks to benchmark performance. However, the above methods assume simple degradation and struggle in blind degradation settings. PGDiff~\cite{yang2023pgdiff} addresses this by leveraging high-quality attribute guidance. Furthermore, SSDiff~\cite{li2025self} guides the sampling process by constructing high-quality pseudo labels for degradation faces. However, the resulting restorations are deterministic and lack diversity. In contrast, our method handles blind degradation while generating diverse, text-aligned outputs through guided sampling.


\section{Preliminaries}
\subsection{Diffusion Models}
The diffusion generative model~\cite{ho2020denoising} is designed to approximate the data distribution $q({\boldsymbol{x}_0})$. It consists of two stages: 

\underline{The Forward Diffusion} process defined as a Markov chain of step length $T$, where Gaussian noise is incrementally added to the data $\boldsymbol{x}_0$ using a fixed noise schedule $\beta_1, ..., \beta_T$, progressively corrupting the original input:
\begin{equation}
q\left( {{\boldsymbol{x}_t}|\boldsymbol{x}_{t-1}} \right) = \mathcal{N}\left( {{\boldsymbol{x}_t};\sqrt {1-{{\beta  }_t}} \boldsymbol{x}_{t-1}, {\beta  }_t \boldsymbol{I}} \right).
\end{equation}
The forward noise processing can be sampled directly from $\boldsymbol{x}_{0}$ to any step $\boldsymbol{x}_t$ by the following equation:
\begin{equation}
{\boldsymbol{x}_t} = \sqrt {{\overline{\alpha}  }_t} \boldsymbol{x}_{0} + \sqrt {1-{\overline{\alpha}  }_t} \boldsymbol{\epsilon}, \quad \boldsymbol{\epsilon} \sim \mathcal{N}\left( {0,\boldsymbol{I}} \right),
\end{equation}
where ${\alpha _t} = 1 - {\beta _t}$ and ${{\overline \alpha }_t} = \prod\limits_{i = 0}^t {{\alpha _i}}$.

\underline{The Reverse Diffusion} process iteratively reconstructs clean data $\boldsymbol{x}_0$ starting from the Gaussian noise distribution $\boldsymbol{x}_T \sim \mathcal{N}\left( {0,\boldsymbol{I}} \right)$, following a Markov chain~\cite{ho2020denoising} structure with Gaussian transitions:
\begin{equation}
p_\theta\left( {{\boldsymbol{x}_{t - 1}}|{\boldsymbol{x}_t}} \right) = \mathcal{N}\left( {{\boldsymbol{x}_{t - 1}};{\boldsymbol{\mu} _\theta}\left( {{\boldsymbol{x}_t},t} \right),\Sigma_\theta  {\left( {{\boldsymbol{x}_t},t} \right)} } \right),
\end{equation}
where $\Sigma_\theta  {\left( {{\boldsymbol{x}_t},t} \right)}  = \frac{{1 - {{\overline \alpha  }_{t - 1}}}}{{1 - {{\overline \alpha  }_t}}}{\beta _t}\boldsymbol{I}$, ${\boldsymbol{\mu} _\theta}\left( {{\boldsymbol{x}_t},t} \right)$ is the mean value we estimate by a deep network $\boldsymbol{\epsilon}_\theta$ from ${\boldsymbol{x}_t}$ as follows:
\begin{equation}
{\boldsymbol{\mu} _\theta}\left( {{\boldsymbol{x}_t}, t} \right) = \frac{1}{{\sqrt {{\alpha _t}} }}\left( {{\boldsymbol{x}_t} - \frac{{\beta _t}}{{\sqrt {1 - {{\overline \alpha  }_t}} }}} \boldsymbol{\epsilon}_\theta \left( {{\boldsymbol{x}_t},t} \right)\right).
\end{equation}
In practice, we can directly approximate ${{\boldsymbol{\hat x}}_0}$ from ${{\boldsymbol{x}}_t}$:
\begin{equation}
{{\boldsymbol{\hat x}}_0} \!=\! {{{\boldsymbol{x}_t}} \mathord{\left/
 {\vphantom {{{\boldsymbol{x}_t}} {\!\!\sqrt {1 \!-\! {{\bar \alpha }_t}} }}} \right.
 \kern-\nulldelimiterspace} {\sqrt {{{\bar \alpha }_t}} }} \!-\! \boldsymbol{\epsilon}_\theta\! \left( {{\boldsymbol{x}_t},t} \right) \!\sqrt {{{\left( {1 \!-\! {{\bar \alpha }_t}} \right)} \mathord{\left/
 {\vphantom {{\left( {1 \!-\! {{\bar \alpha }_t}} \right)} {{{\bar \alpha }_t}}}} \right.
 \kern-\nulldelimiterspace} {{{\bar \alpha }_t}}}}.
\label{eq: approximate}
\end{equation}

\subsection{Inverse Problem Solving With Diffusion Model}
In practice~\cite{kawar2022denoising}, the full data $\boldsymbol{x}_0$ is unavailable; instead, we observe a degraded measurement $\boldsymbol{y}$ through $\boldsymbol{y} = A(\boldsymbol{x}_0) + n$, where $A(\cdot)$ denotes the measurement operator and $n$ is observation noise (\eg, Gaussian noise). 

Diffusion models serve as priors over $p(\boldsymbol{x}_0)$, allowing approximate posterior sampling from $p(\boldsymbol{x}_0|\boldsymbol{y})$ via a modified reverse process. To incorporate measurements into the unconditional diffusion framework, we inject a data-consistency gradient at each step, approximating the likelihood term as:
\begin{equation}
\nabla_{\boldsymbol{x}_t} \log p(\boldsymbol{y} \mid \boldsymbol{x}_t) \approx -\lambda \cdot \nabla_{\boldsymbol{x}_t} \left\| \boldsymbol{y} - A(\hat{\boldsymbol{x}}_0) \right\|^2,
\end{equation}
where $\hat{\boldsymbol{x}}_0$ is an estimate of clean datas derived from $\boldsymbol{x}_t$ using Eq.~(\ref{eq: approximate}). This correction modifies the predicted mean $\tilde{\mu}_\theta$:
\begin{equation}
\tilde{\boldsymbol{\mu}}_\theta(\boldsymbol{x}_t, t, \boldsymbol{y}) = \boldsymbol{\mu}_\theta(\boldsymbol{x}_t, t) - \lambda \cdot \nabla_{\boldsymbol{x}_t} \left\| \boldsymbol{y} - A(\hat{\boldsymbol{x}}_0) \right\|^2,
\end{equation}
used during the posterior-guided reverse sampling step:
\begin{equation}
\boldsymbol{x}_{t-1} \sim \mathcal{N}\left( \tilde{\boldsymbol{\mu}}_\theta(\boldsymbol{x}_t, t, \boldsymbol{y}), \Sigma_\theta \mathbf{I} \right).
\end{equation}
By injecting gradients $\nabla_{\boldsymbol{x}_t} \log p(\boldsymbol{y} \mid \boldsymbol{x}_t)$ into the reverse diffusion steps, inverse problem solving can be performed as approximate posterior sampling~\cite{chung2022diffusion}.

\begin{figure}[t]
\begin{overpic}[width=1\linewidth]{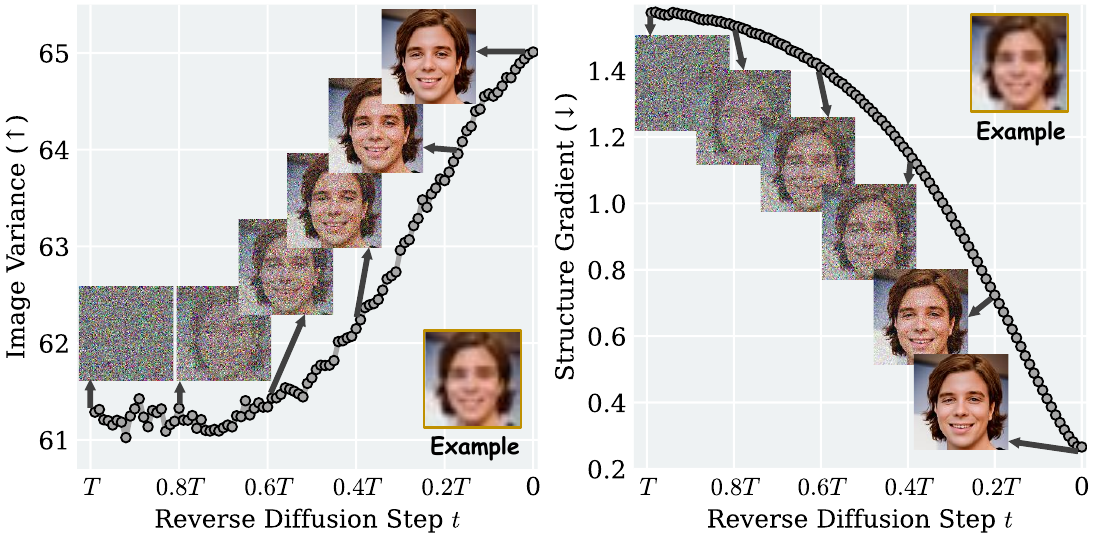}
\end{overpic}
\vspace{-3mm}
   \caption{We perform a statistical analysis of the pre-trained T2I~\cite{saharia2022photorealistic} model's convergence during reverse diffusion on CelebA-HQ test set. Image variance, structural gradients, and selected visualizations all show that the T2I model first reconstructs structural features in a small portion of facial regions, then refines the detailed features covering the majority of facial regions.}
\label{fig: select_reason}
\vspace{-1mm}
\end{figure}

\section{Method}
As shown in Fig.~\ref{fig: pipeline}, for an extremely blurred facial input $\boldsymbol{y}$ and a given textual prompt $\boldsymbol{c}$, our goal is to ensure that restored facial results $\boldsymbol{x}_0$ satisfy both input degradations and prompt conditions. To this end, we reformulate BFR as a balance between degradation consistency and diversity, with degradation consistency enforced by Forward Measurements and diversity achieved through text-guided alignment in inverse projection spaces derived from degradation-aligned sampling. To ensure that Reverse Measurements are applied only after structural features have stabilized, we introduce a Selection Mechanism that determines when each measurement operates. Algorithm~\ref{alg:sampling} summarizes the guided sampling process of our MCS. The following sections detail our guided sampling process, including Forward Measurement, Reverse Measurement, and Selection Mechanism.

\subsection{Selection Mechanism.} \label{subsec:Method}
Before introducing forward and reverse measurements, we first outline the selection mechanism for these approaches at different stages of reverse diffusion. As discussed in Fig.~\ref{fig: insight}, inverse problem solving is applied during reverse diffusion to expand the solution space and integrate text guidance once facial structures stabilize. Figure~\ref{fig: select_reason} presents the convergence trends of pre-trained T2I diffusion models, indicating that reverse diffusion initially focuses on facial structure before gradually refining details. Based on this observation, we apply forward measurement during the initial phase $[T,0.6T]$ to guide facial structure adaptation to degenerate inputs. Once facial structure stabilizes, reverse measurement is introduced during the interval $[0.6T, 0]$, progressively incorporating text alignment constraints to ensure the solution space accommodates varying textual conditions. This coarse-to-fine design ensures that generated images conform to degraded face structures and textual measurements.


\begin{algorithm}[t]
\caption{Sampling process of our MCS.}
\begin{algorithmic}
\STATE \textbf{Input}: Extremely LQ face $\boldsymbol{y}$, text tokens $\boldsymbol{c}$.
\STATE \textbf{Require}: Pre-trained T2I $(\mu_\theta(\boldsymbol{x}_t,t), \Sigma_\theta(\boldsymbol{x}_t,t))$ and Restore $\mathcal{R}$, Degradation $A$ and its pseudo inverse $A^\dagger$.
\STATE \textbf{Output}: Text-aligned HQ face $\boldsymbol{x}_0$.
\STATE $\boldsymbol{x}_t = \sqrt{\bar{\alpha}_t} \cdot \mathcal{R}(\boldsymbol{y}) + \sqrt{1 - \bar{\alpha}_t} \cdot \boldsymbol{\epsilon}, \quad \boldsymbol{\epsilon} \sim \mathcal{N}(0, \mathbf{I})$
\FOR{$t$ from $T$ to $1$}
    \STATE ${\hat {\boldsymbol{x}}_t} \leftarrow \frac{1}{{\sqrt {\bar{\alpha}_t}}} \boldsymbol{x}_t - \sqrt{\frac{1 - \bar{\alpha}_t}{\bar{\alpha}_t}} \boldsymbol{\varepsilon}_\theta(\boldsymbol{x}_t, t, \boldsymbol{c})$
    \IF{$t$ in Selection}
        \STATE $\boldsymbol{x}_{t-1} \sim \boldsymbol{x}_{t} \!-\! \nabla_{\hat{\boldsymbol{x}}_t} \| \mathcal{R}(\boldsymbol{y})|_{Struct} \!-\! \hat{\boldsymbol{x}}_t|_{Struct} \|_2^2$
    \ELSE
        \STATE $\boldsymbol{x}_{t-1} \sim \boldsymbol{x}_{t} \!-\! \nabla_{\hat{\boldsymbol{x}}_t} \| \mathcal{R}(\boldsymbol{y}) \!-\! A^\dagger\left(A(\hat{\boldsymbol{x}}_t)\right)\!\|_2^2$
    \ENDIF
    \STATE $\boldsymbol{x}_{t-1} \sim \mathcal{N}(\mu_\theta(\boldsymbol{x}_t,t) - \Sigma_\theta(\boldsymbol{x}_t,t) \nabla \mathcal{L}, \Sigma_\theta(\boldsymbol{x}_t,t))$
\ENDFOR
\RETURN $\boldsymbol{x}_0$
\end{algorithmic}
\label{alg:sampling}
\end{algorithm}

\subsection{Forward Measurement.} 
As shown in Fig.~\ref{fig: pipeline}, forward measurement is formulated as a joint guidance of textual prompts and coarse HR estimates. First, let $\boldsymbol{p}$ denote a user-specified textual prompt (\eg, “a woman with glasses”), which is embedded by a pretrained text tokenizer ${\boldsymbol{\varepsilon}_{\text{text}}}$ in the T2I Diffusion model~\cite{saharia2022photorealistic} to produce contextual features $\boldsymbol{c} = {\boldsymbol{\varepsilon}_{\text{text}}}(\boldsymbol{p})$. These embeddings are injected into each denoising step of the UNet backbone within the diffusion model, thereby conditioning the predicted posterior distribution:
\begin{equation}
p_\theta(\boldsymbol{x}_{t-1} \mid \boldsymbol{x}_t, \boldsymbol{c}) = \mathcal{N}(\boldsymbol{\mu}_\theta(\boldsymbol{x}_t, t, \boldsymbol{c}), \Sigma_\theta \mathbf{I}),
\end{equation}
where ${\boldsymbol{\mu}_\theta}$ is the model-predicted mean conditioned on the noisy input $\boldsymbol{x}_t$ and prompt embedding $\boldsymbol{c}$, guiding the generative trajectory toward the target semantics. However, relying solely on textual constraints often fails to produce results closely aligned with the desired output.

\begin{table*}
\setlength\tabcolsep{2pt}
  \centering
  \caption{Quantitative comparisons on the synthetic CelebA-HQ~\cite{karras2017progressive} dataset. \textbf{Bold} and \underline{underline} indicates the best and the second best performance. ``L'' and ``S'' represent ``Learning-based (Paired-data Training)'' and ``Sampling-based (Zero-shot, Train-free)''. For this Table and Table~\ref{tab:real_performance}, our method is evaluated under the prompt-free setting (\ie, prompts are set to empty).}
  \label{tab:syn_performance}
  \vspace{-2mm}
  \resizebox{0.9999\textwidth}{!}{
      \begin{tabular}{c|c|c c c c|c c c c|c c c c}
        \toprule
         & & \multicolumn{4}{c|}{$\times16$} & \multicolumn{4}{c|}{$\times8$} & \multicolumn{4}{c}{$\times4$} \\
         \multicolumn{1}{c|}{\multirow{-2}{*}{Methods}} & \multicolumn{1}{c|}{\multirow{-2}{*}{Type}} & MUSIQ$\uparrow$ & MAN-IQA$\uparrow$ & DISTS$\downarrow$ & LPIPS$\downarrow$ & MUSIQ$\uparrow$ & MAN-IQA$\uparrow$ & DISTS$\downarrow$ & LPIPS$\downarrow$ & MUSIQ$\uparrow$ & MAN-IQA$\uparrow$ & DISTS$\downarrow$ & LPIPS$\downarrow$ \\
        \midrule
         DR2~\cite{wang2023dr2} & L  & 27.05 & 0.2225  & 0.3210 & 0.4785 & 31.72 & 0.2483 & 0.2899 & 0.4224 & 21.81 & 0.1696 & 0.3280 & 0.4701 \\
         DifFace~\cite{yue2024difface} & L  & 37.77 & 0.2042  & 0.2657 & 0.3824 & 54.51 & 0.2870 & 0.2239 & 0.2966 & 61.38 & 0.3574 & 0.1955 & 0.2431 \\
         DiffBIR~\cite{lin2024diffbir} & L  & 60.45 & 0.3893  & \textbf{0.2104} & \textbf{0.2856} & 61.98 & \underline{0.3926} & \textbf{0.1957} & \textbf{0.2489} & 61.00 & \underline{0.3808} & \underline{0.1855} & \textbf{0.2302} \\
        \midrule
        DPS~\cite{chung2022diffusion}  & S  & 54.36 & \underline{0.3896}  & 0.3171 & 0.4917 & 41.74 & 0.3515 & 0.3416 & 0.4973 & 26.51 & 0.1856 & 0.3249 & 0.5058 \\
        IIGDM~\cite{song2023pseudoinverse}  & S  & 49.89 & 0.3375  & 0.3283 & 0.5037 & 37.88 & 0.3002 & 0.3624 & 0.5328 & 28.79 & 0.2828 & 0.3916 & 0.5630 \\
        DDNM~\cite{wang2022zero} & S  & 29.30 & 0.1994  & 0.3817 & 0.4479 & 24.23 & 0.2091 & 0.3370 & 0.5010 & 29.29 & 0.2372 & 0.3133 & 0.4479 \\
        PGDiff~\cite{yang2023pgdiff}  & S & 55.36 & 0.3620  & 0.2193 & 0.3231 & 57.27 & 0.3717 & 0.2043 & 0.2784 & 56.18 & 0.3771 & 0.1986 & 0.2644  \\
        MCG-Diff~\cite{cardoso2023monte} & S & 45.28  & 0.3497  & 0.3466 & 0.4734 & 38.45 & 0.3342 & 0.3359 & 0.4972 & 27.89 & 0.2431 & 0.3298 & 0.5163 \\
        SSDiff~\cite{li2025self} & S & \underline{61.05}  & 0.3753  & 0.2228 & 0.3129 & \underline{62.22} & 0.3771 & 0.2021 & 0.2713 & \underline{63.88} & 0.3750 & 0.1882 & 0.2459 \\
        \midrule
         \textbf{Ours} & S & \textbf{62.72} & \textbf{0.4483}  & \underline{0.2215}  & \underline{0.3115} & \textbf{63.02} & \textbf{0.4436} & \underline{0.1983} & \underline{0.2649} & \textbf{64.82} & \textbf{0.4632} & \textbf{0.1851} & \underline{0.2361} \\
        \bottomrule
      \end{tabular}
  }
  \vspace{-2mm}
\end{table*}

\begin{table}
\setlength\tabcolsep{2.5pt}
  \centering
  \caption{Quantitative comparisons on real-world datasets~\cite{wang2021towards,zhou2022towards}. \textbf{Bold} and \underline{underline} indicates the best and the second best results. ``L'' and ``S'' represent ``Learning-based (Paired-data Training)'' and ``Sampling-based (Zero-shot, Train-free)''.}
  \label{tab:real_performance}
  \vspace{-0mm}
  \resizebox{0.48\textwidth}{!}{
      \begin{tabular}{c|c|c c|c c}
        \toprule
         & & \multicolumn{2}{c|}{Wider-Test~\cite{zhou2022towards}} & \multicolumn{2}{c}{WebPhoto-Test~\cite{wang2021towards}}  \\
         \multicolumn{1}{c|}{\multirow{-2}{*}{Methods}} & \multicolumn{1}{c|}{\multirow{-2}{*}{Type}} & MUSIQ$\uparrow$ & MAN-IQA$\uparrow$  & MUSIQ$\uparrow$ & MAN-IQA$\uparrow$ \\
        \midrule
         DR2~\cite{wang2023dr2} & L  & 39.27 & 0.2602   & 32.27 & 0.2632  \\
         DifFace~\cite{yue2024difface} & L  & 60.38 & 0.3516   & 59.50  & 0.3365  \\
         DiffBIR~\cite{lin2024diffbir} & L  & \underline{62.57} & 0.3613  & 65.62 & \underline{0.4009}  \\
        \midrule
        DPS~\cite{chung2022diffusion}  & S  & 23.28 & 0.1571  & 21.22 & 0.2632  \\
        IIGDM~\cite{song2023pseudoinverse}  & S  & 22.67 & 0.1557  & 42.91 & 0.2831  \\
        DDNM~\cite{wang2022zero} & S  & 22.80 &  0.1959  & 22.13 & 0.1485  \\
        PGDiff~\cite{yang2023pgdiff}  & S & 61.28 & 0.3496  & 61.75 & 0.3655    \\
        MCG-Diff~\cite{cardoso2023monte} & S & 23.34 & 0.1533  & 20.57 & 0.2493  \\
        SSDiff~\cite{li2025self} & S & 62.44 & \underline{0.3669}  & \underline{66.17} & 0.4002  \\
        \midrule
         \textbf{Ours} & S & \textbf{68.08} & \textbf{0.4236}  & \textbf{69.74} & \textbf{0.4652} \\
        \bottomrule
      \end{tabular}
  }
  \vspace{-2mm}
\end{table}

To incorporate semantic constraints, we initialize the sampling at timestep $t$ with a coarse face estimate $\boldsymbol{y}_0$ obtained from a pretrained restorer~\cite{yang2021gan}, and perturb it with Gaussian noise to align with the diffusion distribution:
\begin{equation}
\boldsymbol{x}_t = \sqrt{\bar{\alpha}_t} \cdot \boldsymbol{y}_0 + \sqrt{1 - \bar{\alpha}_t} \cdot \boldsymbol{\epsilon}, \quad \boldsymbol{\epsilon} \sim \mathcal{N}(0, \mathbf{I}).
\end{equation}
Unlike previous works~\cite{wang2023dr2,lin2024diffbir} that initialize from pure noise, we ensure that early samples are structurally consistent with high-quality attribute observation~\cite{yang2023pgdiff,li2025self}. Specifically, at each denoising step, denoise models predict a clean signal $\hat{\boldsymbol{x}}_t = \boldsymbol{\mu}_\theta(\boldsymbol{x}_t, t, \boldsymbol{c})$, conditioned on prompt semantics. To preserve consistency with the proxy $\boldsymbol{y}_0$, we first apply Haar wavelet decomposition $\mathcal{WD}$, thereby obtaining a combination comprising a high-frequency portion $V\!H\!D$ formed by merging three high-frequency subbands and a low-frequency subband portion $L$. To facilitate the convergence of structures in intermediate states, we select high-frequency portion that primarily reflect facial structure:
\begin{equation}
\left\{ {\boldsymbol{y}_L^0,\boldsymbol{y}_{V\!H\!D}^0} \right\},\left\{ {\hat{\boldsymbol{x}}_L^t,\hat{\boldsymbol{x}}_{V\!H\!D}^t} \right\} \!=\! \mathcal{WD}\left( {{\boldsymbol{y}_0}} \right),\mathcal{WD}\left( {{\hat{\boldsymbol{x}}_t}} \right).
\end{equation}
Then we impose a forward measurement loss:
\begin{equation}
\mathcal{L}_1 = \left\| \boldsymbol{y}_{V\!H\!D}^0 - \hat{\boldsymbol{x}}_{V\!H\!D}^t \right\|_2^2,
\end{equation}
Meanwhile, the prompt embedding $\boldsymbol{c}$ affects the gradient through $\hat{\boldsymbol{x}}_t$, allowing semantic cues (\eg, “with glasses”) to influence the direction and content of the update.

Finally, the next intermediate step $\boldsymbol{x}_{t-1}$ is refined through a gradient-guided sampling scheme~\cite{dhariwal2021diffusion}:
\begin{equation}
\boldsymbol{x}_{t-1} \leftarrow \sqrt{\bar{\alpha}_t} \cdot \boldsymbol{y}_0 \!+\! \sqrt{1 - \bar{\alpha}_t} \cdot \boldsymbol{\epsilon} \!-\! \eta\nabla_{\boldsymbol{x}_t} \mathcal{L}_1 + \sigma_t \cdot \boldsymbol{\epsilon},
\end{equation}
where $\eta$ is a step size hyperparameter. This update encourages the generation to remain faithful to facial measurement structure, ensuring that input degenerate attributes (\eg, overall facial structures of inputs) manifest coherently in outputs.

\begin{figure}[t]
\begin{overpic}[width=0.99\linewidth]{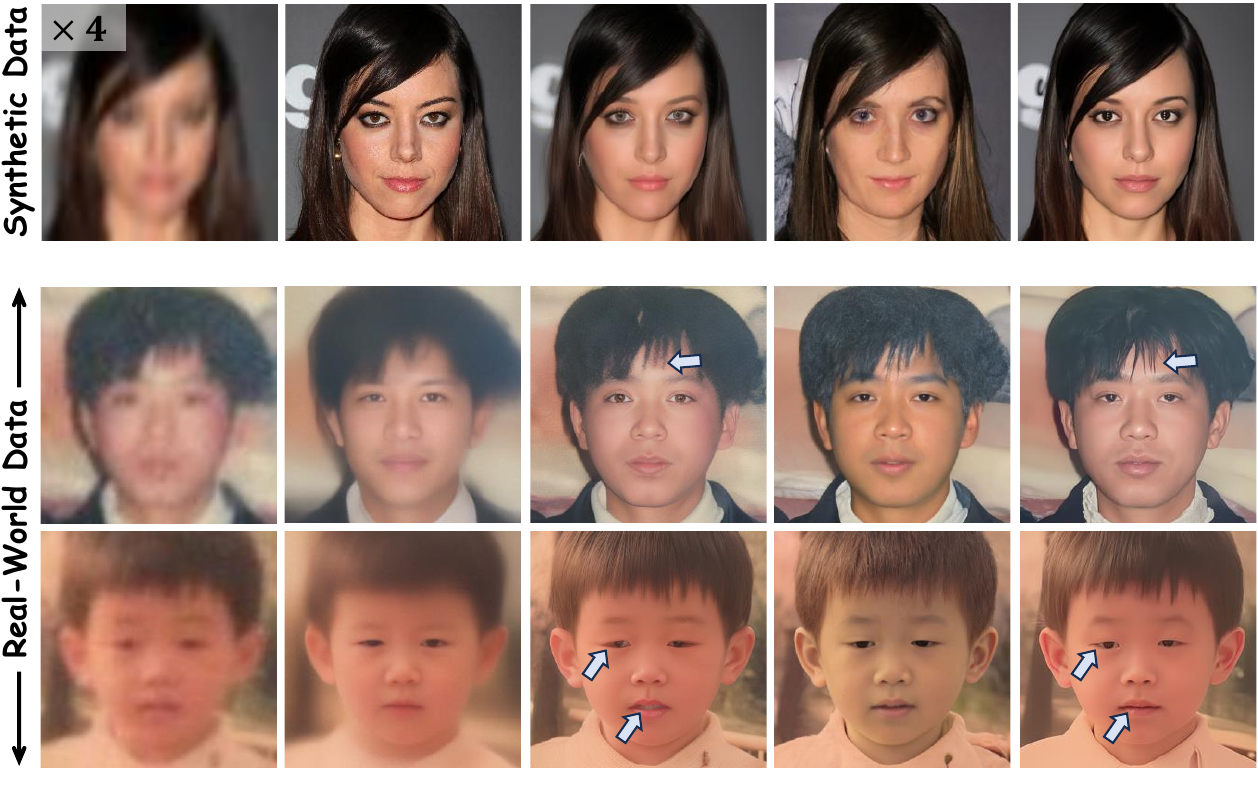}

\put(8.0,0.3){\color{black}{\fontsize{7.5pt}{1pt}\selectfont Real LR}}
\put(25.8,0.3){\color{black}{\fontsize{7.5pt}{1pt}\selectfont DR2~\cite{wang2023dr2}}}
\put(43.2,0.3){\color{black}{\fontsize{7.5pt}{1pt}\selectfont DiffBIR~\cite{lin2024diffbir}}}
\put(63.6,0.3){\color{black}{\fontsize{7.5pt}{1pt}\selectfont SSDiff~\cite{li2025self}}}
\put(86.9,0.3){\color{black}{\fontsize{7.5pt}{1pt}\selectfont \textbf{Ours}}}

\put(4.0,42.6){\color{black}{\fontsize{7.5pt}{1pt}\selectfont Synthetic LR}}
\put(29.8,42.6){\color{black}{\fontsize{7.5pt}{1pt}\selectfont GT}}
\put(43.2,42.6){\color{black}{\fontsize{7.5pt}{1pt}\selectfont DiffBIR~\cite{lin2024diffbir}}}
\put(63.6,42.6){\color{black}{\fontsize{7.5pt}{1pt}\selectfont SSDiff~\cite{li2025self}}}
\put(86.9,42.6){\color{black}{\fontsize{7.5pt}{1pt}\selectfont \textbf{Ours}}}
\end{overpic}
\vspace{0mm}
   \caption{Quantitative results under mild degradation without textual prompts. Our method can achieve higher BFR quality than the latest learning-based~\cite{wang2023dr2,lin2024diffbir} and sampling-based~\cite{li2025self} methods.}
\label{fig: small_scale}
\end{figure}
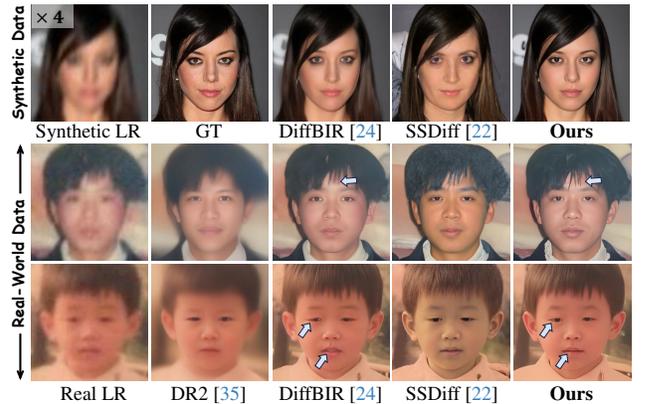

\subsection{Reverse Measurement.} 
Despite the inclusion of textual prompts in the forward measurement, due to the deterministic nature of forward measurements, overly strong constraints may suppress the diversity of samples aligned with textual prompts. To balance measurement fidelity and semantic expressiveness, as shown in Fig.~\ref{fig: pipeline}, we introduce an inverse measurement constraint. This constraint encourages generated contents to align with degraded inputs while expanding the potential generative space. Centered on the constructed anchor $A(\boldsymbol{y}_0)$ with a pseudo-inverse range $A^\dagger$ as its radius, this expansion ensures that results remain consistent with the given textual prompt.

\begin{table*}[t!]
\tiny
\centering
\setlength\tabcolsep{3pt}
\caption{Comparison on face landmark distance (LMD) ($\downarrow$) to evaluate fidelity on $\times8$ CelebA-HQ. \emph{Following previous works~\cite{wang2025osdface,li2025self}, we don't employ PSNR/SSIM to evaluate fidelity, as these metrics primarily reflect pixel consistency rather than facial identity preservation.}}
\vspace{-2mm}
\centering
\resizebox{0.99\textwidth}{!}{
\begin{tabular}{@{}c|cccccccccc@{}}
\toprule
Input &DR2~\cite{wang2023dr2} &DifFace~\cite{yue2024difface} &DiffBIR~\cite{lin2024diffbir} &DPS~\cite{chung2022diffusion}	& IIGDM~\cite{song2023pseudoinverse} &DDNM~\cite{wang2022zero} &PGDiff~\cite{yang2023pgdiff} &MCG-Diff~\cite{cardoso2023monte} &SSDiff~\cite{li2025self} & \textbf{Ours}\\
\hline
180.4  & 3.165 & 2.874 &\textbf{2.056} &5.406 &6.791	&3.010 & 3.031 & 5.742 & 2.413 & \underline{2.268}\\
\toprule
\end{tabular}}
\vspace{-1mm}
\label{table: identity_distances}
\end{table*}

\begin{figure*}[t]
\begin{overpic}[width=0.99\linewidth]{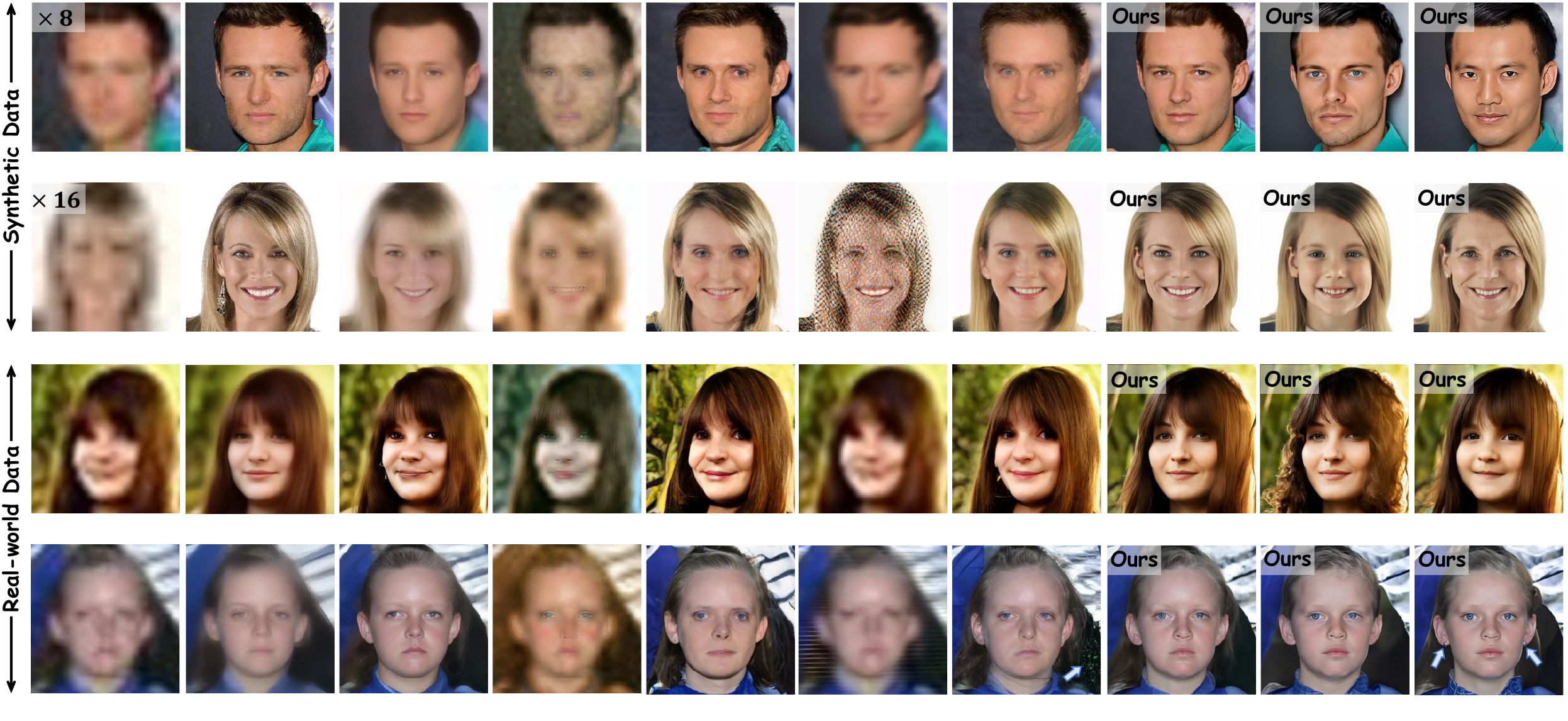}
\put(4.0,0.3){\color{black}{\fontsize{7.5pt}{1pt}\selectfont Real LR}}
\put(14.1,0.3){\color{black}{\fontsize{7.5pt}{1pt}\selectfont DR2~\cite{wang2023dr2}}}
\put(22.5,0.3){\color{black}{\fontsize{7.5pt}{1pt}\selectfont DiffBIR~\cite{lin2024diffbir}}}
\put(32.3,0.3){\color{black}{\fontsize{7.5pt}{1pt}\selectfont DDNM~\cite{wang2022zero}}}
\put(42.2,0.3){\color{black}{\fontsize{7.5pt}{1pt}\selectfont PGDiff~\cite{yang2023pgdiff}}}
\put(51.5,0.3){\color{black}{\fontsize{7.5pt}{1pt}\selectfont MCG-Diff~\cite{cardoso2023monte}}}
\put(62.0,0.3){\color{black}{\fontsize{7.5pt}{1pt}\selectfont SSDiff~\cite{li2025self}}}
\put(73.3,0.3){\color{black}{\fontsize{7.5pt}{1pt}\selectfont ``Null''}}
\put(81.5,0.3){\color{black}{\fontsize{7.5pt}{1pt}\selectfont ``Little boy''}}
\put(90.2,0.3){\color{black}{\fontsize{7.5pt}{1pt}\selectfont ``With earrings''}}

\put(4.0,11.8){\color{black}{\fontsize{7.5pt}{1pt}\selectfont Real LR}}
\put(14.1,11.8){\color{black}{\fontsize{7.5pt}{1pt}\selectfont DR2~\cite{wang2023dr2}}}
\put(22.5,11.8){\color{black}{\fontsize{7.5pt}{1pt}\selectfont DiffBIR~\cite{lin2024diffbir}}}
\put(32.3,11.8){\color{black}{\fontsize{7.5pt}{1pt}\selectfont DDNM~\cite{wang2022zero}}}
\put(42.2,11.8){\color{black}{\fontsize{7.5pt}{1pt}\selectfont PGDiff~\cite{yang2023pgdiff}}}
\put(51.5,11.8){\color{black}{\fontsize{7.5pt}{1pt}\selectfont MCG-Diff~\cite{cardoso2023monte}}}
\put(62.0,11.8){\color{black}{\fontsize{7.5pt}{1pt}\selectfont SSDiff~\cite{li2025self}}}
\put(73.3,11.8){\color{black}{\fontsize{7.5pt}{1pt}\selectfont ``Null''}}
\put(81.3,11.8){\color{black}{\fontsize{7.5pt}{1pt}\selectfont ``Curly hair''}}
\put(92.5,11.8){\color{black}{\fontsize{7.5pt}{1pt}\selectfont ``Child''}}

\put(2.5,23.4){\color{black}{\fontsize{7.5pt}{1pt}\selectfont Synthetic LR}}
\put(15.3,23.4){\color{black}{\fontsize{7.5pt}{1pt}\selectfont GT}}
\put(23.1,23.4){\color{black}{\fontsize{7.5pt}{1pt}\selectfont DR2~\cite{wang2023dr2}}}
\put(32.3,23.4){\color{black}{\fontsize{7.5pt}{1pt}\selectfont DDNM~\cite{wang2022zero}}}
\put(42.2,23.4){\color{black}{\fontsize{7.5pt}{1pt}\selectfont PGDiff~\cite{yang2023pgdiff}}}
\put(51.5,23.4){\color{black}{\fontsize{7.5pt}{1pt}\selectfont MCG-Diff~\cite{cardoso2023monte}}}
\put(62.0,23.4){\color{black}{\fontsize{7.5pt}{1pt}\selectfont SSDiff~\cite{li2025self}}}
\put(73.3,23.4){\color{black}{\fontsize{7.5pt}{1pt}\selectfont ``Null''}}
\put(81.3,23.4){\color{black}{\fontsize{7.5pt}{1pt}\selectfont ``Little girl''}}
\put(90.3,23.4){\color{black}{\fontsize{7.5pt}{1pt}\selectfont ``Middle-aged''}}

\put(2.5,35.0){\color{black}{\fontsize{7.5pt}{1pt}\selectfont Synthetic LR}}
\put(15.3,35.0){\color{black}{\fontsize{7.5pt}{1pt}\selectfont GT}}
\put(23.1,35.0){\color{black}{\fontsize{7.5pt}{1pt}\selectfont DR2~\cite{wang2023dr2}}}
\put(32.3,35.0){\color{black}{\fontsize{7.5pt}{1pt}\selectfont DDNM~\cite{wang2022zero}}}
\put(42.2,35.0){\color{black}{\fontsize{7.5pt}{1pt}\selectfont PGDiff~\cite{yang2023pgdiff}}}
\put(51.5,35.0){\color{black}{\fontsize{7.5pt}{1pt}\selectfont MCG-Diff~\cite{cardoso2023monte}}}
\put(62.0,35.0){\color{black}{\fontsize{7.5pt}{1pt}\selectfont SSDiff~\cite{li2025self}}}
\put(73.3,35.0){\color{black}{\fontsize{7.5pt}{1pt}\selectfont ``Null''}}
\put(80.5,35.0){\color{black}{\fontsize{7.5pt}{1pt}\selectfont ``Angular face''}}
\put(90.1,35.0){\color{black}{\fontsize{7.5pt}{1pt}\selectfont ``Asian-looking''}}

\end{overpic}
\vspace{-0mm}
   \caption{Our method maintains high-fidelity restoration under extremely LR conditions, producing visually coherent results when prompts are not provided. Furthermore, our method can achieve accurate semantic alignment when guided by text prompts.}
\label{fig: big_scale}
\end{figure*}
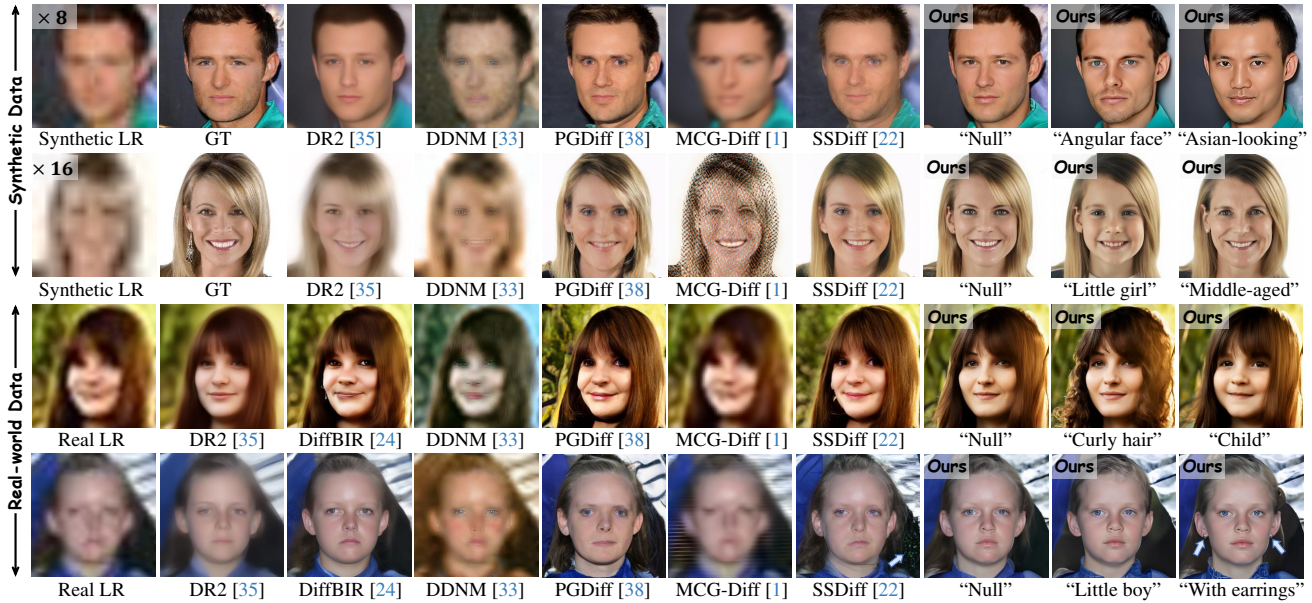

Specifically, we define a known degradation operator \( A \) (\eg, average pooling) that maps the coarse HR estimate face $\boldsymbol{y}_0$ to the observation anchor $A(\boldsymbol{y}_0)$, and solve its pseudo-inverse \( A^\dagger \) using the Singular Value Decomposition (SVD) following~\cite{li2019controllable,kawar2022denoising}, which approximates the inverse mapping. At each denoising step, similar to operators in Forward Measurement, we first obtain the predicted clean signal \( \hat{\boldsymbol{x}}_t = \mu_\theta(\boldsymbol{x}_t, t, \boldsymbol{c}) \). Then we solve the inverse problem on \( \hat{\boldsymbol{x}}_t \) to encourage alignment with the degraded observation:
\begin{equation}
\mathcal{L}_2 = \left\| \boldsymbol{y}_0 - A^\dagger\left(A(\hat{\boldsymbol{x}}_t)\right) \right\|_2^2.
\end{equation}
This measurement is centered on the anchor $A(\boldsymbol{y}_0)$. Geometrically, the reverse measurement constrains $\hat{\boldsymbol{x}}_t$ within the solution space and centres the space on the degenerate anchor point $A(\boldsymbol{y}_0)$. The pseudo-inverse projection via $A^\dagger$ steers the reconstruction toward a solution similar to $\boldsymbol{y}_0$. As the state of $\hat{\boldsymbol{x}}_t$ fluctuates at each step during reverse diffusion, $AA^\dagger$ enables the generation of a solution space, which is then guided towards consistent solutions through prompts.

Compared to minimizing $\left\|A(\boldsymbol{y}_0) - A(\hat{\boldsymbol{x}}_t)\right\|_2^2$~\cite{chung2022diffusion}, which enforces a more strict match in observation spaces, our formulation with $A^\dagger A$ introduces a relaxed constraint, allowing for greater diversity. Then, $\boldsymbol{x}_{t-1}$ is refined through gradient-guided sampling. The process is as follows:
\begin{equation}
\boldsymbol{x}_{t-1} \leftarrow \sqrt{\bar{\alpha}_t} \cdot \boldsymbol{y}_0 \!+\! \sqrt{1 - \bar{\alpha}_t} \cdot \boldsymbol{\epsilon} \!-\! \eta\nabla_{\boldsymbol{x}_t} \mathcal{L}_2 \!+\! \sigma_t \cdot \boldsymbol{\epsilon}.
\end{equation}
where $\eta$ is a step size hyperparameter. This update mechanism encourages the generation to adhere faithfully to given textual conditions (\eg, ``a woman with glasses'').

Overall, by staged selection optimization of gradient-based loss \(\mathcal{L}_1\) and \(\mathcal{L}_2\), the T2I diffusion achieves a trade-off between degenerate measurement consistency and textual measurement consistency, enabling generated content to both reflect textual descriptions and satisfy degenerate inputs.

\section{Experiments}
\subsection{Datasets}
Our method is training-free and does not require any training dataset. For the synthetic test set, we follow previous BFR methods~\cite{chan2021glean,he2022gcfsr,chan2022glean} by selecting 1000 HQ face images from CelebA-HQ~\cite{karras2017progressive}, resizing these images to the size of ${256^2}$~\cite{wang2022zero,ding2024restoration,song2023pseudoinverse} as ground truth ${\boldsymbol{I}_{GT}}$. Following the established protocol~\cite{wang2021towards}, we synthesize the corresponding LQ images ${\boldsymbol{I}_{LQ}}$ using the degradation model described below:
\begin{equation}
{\boldsymbol{I}_{LQ}} = {\left[ {\left( {{\boldsymbol{I}_{GT}} \circledast {\boldsymbol{k}_\sigma }} \right){ \downarrow _s} + {\boldsymbol{n}_\delta }} \right]_{\boldsymbol{JPEG}_q}}.
\end{equation}
For the Gaussian blur kernel ${\boldsymbol{k}_\sigma }$, downsampling operation with a scale factor $s$ ($s$ = 8,16,32), Gaussian noise ${\boldsymbol{n}_\delta }$, and JPEG compressing with quality factor $q$, we randomly sample $\sigma$, $\delta$, $q$ from $\left\{ {0.2:10} \right\}$, $\left\{ {0:15} \right\}$, $\left\{ {60:100} \right\}$, respectively. For real-world test sets, we utilize two widely-used real-world face benchmarks with degradation levels ranging from heavy to medium: Wider-Test (970 images)~\cite{zhou2022towards} and Webphoto-Test (407 images)~\cite{wang2021towards}.

\subsection{Implementations}
For pre-trained Text2Image diffusion models, we use the open-source version of Imagen Deep-Floyd IF~\cite{saharia2022photorealistic}. The total number of iteration steps is defaulted to 150. Previous methods, including DifFace~\cite{yue2024difface}, DiffBIR~\cite{lin2024diffbir}, PGDiff~\cite{yang2023pgdiff}, and SSDiff~\cite{li2025self}, are implemented at a resolution of ${512^2}$. However, due to the resolution limitations of the pre-trained diffusion model~\cite{saharia2022photorealistic} used in our approach, following the strategy of previous pre-trained diffusion-based methods~\cite{wang2022zero,ding2024restoration,song2023pseudoinverse,cardoso2023monte}, we use official pre-trained weights of ${512^2}$ resolution-based methods to infer outputs and resize these from ${512^2}$ to ${256^2}$ to ensure a fair comparison. As for assessment metrics, we utilize referenced metrics, including DISTS~\cite{ding2020image}, LPIPS~\cite{zhang2018unreasonable}, and LMD~\cite{gu2022vqfr}. \emph{PSNR and SSIM are not used as they primarily measure pixel similarity rather than face structural similarity.} Non-referenced metrics, including MUSIQ~\cite{ke2021musiq} and MAN-IQA~\cite{yang2022maniqa}. The degradation observation radius $A$ for synthetic data is consistent with the super-resolution scale. For real data, we uniformly set the observation radius to 8. All experiments are conducted with PyTorch on an NVIDIA GeForce RTX 4090.

\begin{figure}[t]
\begin{overpic}[width=0.999\linewidth]{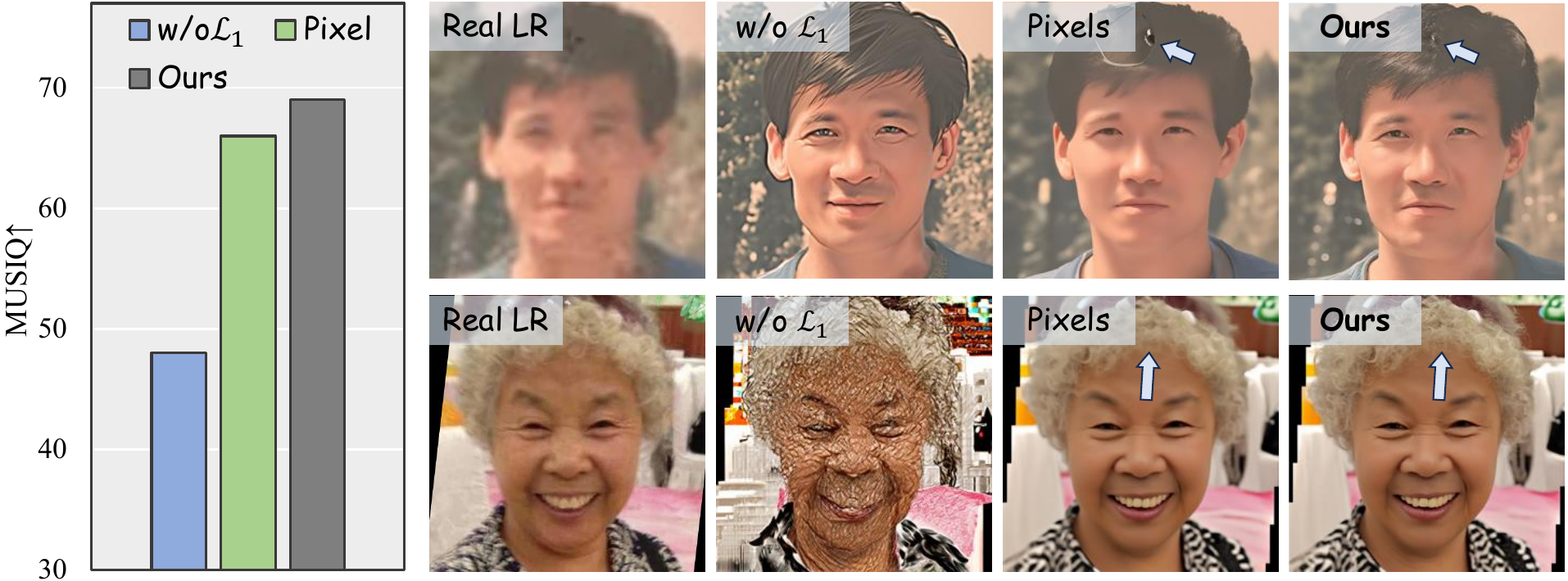}
\end{overpic}
\vspace{-4mm}
   \caption{Ablation studies on our forward measurement ($\mathcal{L}_1$), where ``Pixels'' denotes per-pixel gradient-based descent rather than facial structure-based descent like ours.}
\label{fig: forward_measure}
\vspace{-1mm}
\end{figure}

\vspace{-4mm}
\paragraph{Comparison with Existing Methods}
We compare our method with two types of existing diffusion-based BFR methods, including: i) Learning-based methods that need paired-data training, including DR2~\cite{wang2023dr2}, DifFace~\cite{yue2024difface}, and DiffBIR~\cite{lin2024diffbir}. ii) Sampling-based methods that are zero-shot and train-free, including DPS~\cite{chung2022diffusion}, IIGDM~\cite{song2023pseudoinverse}, DDNM~\cite{wang2022zero}, PGDiff~\cite{yang2023pgdiff}, MCG-Diff~\cite{cardoso2023monte}, and SSDiff~\cite{li2025self}. 

\vspace{-4mm}
\paragraph{Quantitative Comparison.} In Table~\ref{tab:syn_performance} and Table~\ref{tab:real_performance}, we report results on synthetic~\cite{karras2017progressive} and real test sets~\cite{wang2021towards,zhou2022towards}. Our method achieves the best performance across all no-reference metrics, demonstrating its superiority in generating visually high-quality face images. For full-reference metrics, our approach is only surpassed by DiffBIR~\cite{lin2024diffbir} in a subset of metrics, further indicating competitiveness. Furthermore, in TABLE.~\ref{table: identity_distances}, following previous works~\cite{wang2025osdface,li2025self}, we report face landmark distance (LMD) on CelebA-HQ ($\times8$) to measure face fidelity; our method is second only to DiffBIR~\cite{lin2024diffbir}. These results fully demonstrate that our MCS achieves an excellent balance between restoration fidelity and quality.

\vspace{-4mm}
\paragraph{Qualitative Comparison.} As shown in Fig.~\ref{fig: small_scale}, we present visual results on synthetic data of ×4 scale and real-world data under mild degradation. On synthetic data, our results are perceptually closer to GT, while on real data, our method restores realistic facial details. Since mild degradation lacks strong one-to-many ambiguity, text-guided results are omitted. In Fig.~\ref{fig: big_scale}, we evaluate on synthetic data with severe degradations (\eg, ×8 and ×16) and real data. Our method consistently recovers GT-aligned faces on synthetic data and produces natural faces on real data. Moreover, our method effectively incorporates textual prompts, producing results that are not only visually coherent but also aligned with texts, covering attributes such as facial features, age, and accessories, showing the potential of our framework. \textbf{\emph{More poses of LQ faces can be seen in the supplementary.}}

\begin{table}[t!]
\tiny
\setlength\tabcolsep{2.5pt}
\centering
\vspace{0mm}
    \caption{Ablation on text response rate (Reply) and the quality of reconstructed faces (MUSIQ) on $\times 8$ CelebA-HQ, where DDNM~\cite{wang2022zero}, DPS~\cite{chung2022diffusion}, and IIGDM~\cite{song2023pseudoinverse} are under known degradation.}
\vspace{-2mm}
\label{tab: reply}
\resizebox{0.475\textwidth}{!}{
\begin{tabular}{l|c|c|c|c|c}
\toprule
    \multicolumn{1}{l|}{\multirow{-1}{*}{Method}} 
    & \multicolumn{1}{c|}{\multirow{-1}{*}{w/o $\mathcal{L}_2$}}
    & \multicolumn{1}{c|}{\multirow{-1}{*}{DPS~\cite{chung2022diffusion}}}
    & \multicolumn{1}{c|}{\multirow{-1}{*}{DDNM~\cite{wang2022zero}}}
    & \multicolumn{1}{c|}{\multirow{-1}{*}{IIGDM~\cite{song2023pseudoinverse}}}
    & \multicolumn{1}{c}{\multirow{-1}{*}{\textbf{Ours}}}

    \\ 
    \hline
    Reply $\uparrow$    & 41\%   & 32\% & 88\%    & 36\% & \textbf{91\%} \\ 
    MUSIQ $\uparrow$    & 53.4   & 50.1 & 22.1    & 52.2 & \textbf{63.0} \\ 

    \toprule
\end{tabular}}
\end{table}

\begin{table}[t!]
\tiny
\setlength\tabcolsep{4.5pt}
\centering
\vspace{0mm}
    \caption{Ablation on \(\mathcal{L}_1\)/\(\mathcal{L}_2\) Weight Ratio: balance on text response rate (Reply) and face quality (MUSIQ).}
\vspace{-2mm}
\label{tab: Ratio}
\resizebox{0.475\textwidth}{!}{
\begin{tabular}{c|c|c|c|c|c}
\toprule
    \multicolumn{1}{l|}{\multirow{-1}{*}{Weight Ratio}} 
    & \multicolumn{1}{c|}{\multirow{-1}{*}{2/1}}
    & \multicolumn{1}{c|}{\multirow{-1}{*}{1.5/1}}
    & \multicolumn{1}{c|}{\multirow{-1}{*}{\textbf{1/1 (Ours)}}}
    & \multicolumn{1}{c|}{\multirow{-1}{*}{1/1.5}}
    & \multicolumn{1}{c}{\multirow{-1}{*}{1/2}}

    \\ 
    \hline
    Reply $\uparrow$    & 64\%   & 78\% & 91\%    & 93\% & \textbf{93\%} \\ 
    MUSIQ $\uparrow$    & 59.5   & 59.2 & \textbf{63.0}    & 57.5 & 53.3 \\ 

    \toprule
\end{tabular}}
\end{table}

\subsection{Ablation Studies}
\paragraph{Effectiveness of Forward Measurement.}
As shown in Fig.~\ref{fig: forward_measure}, we show the effectiveness of our forward measurement. By injecting guidance from a coarse restore estimate, it mitigates early-stage uncertainty and artifacts caused by noise-dominated predictions. In addition, it constrains the inverse measurement process by suppressing excessive uncertainty that may deviate from the true solution. Compared to per-pixel gradient-based descent, it also improves the perceptual quality, as reflected by metrics such as MUSIQ.

\begin{figure}[t]
\begin{overpic}[width=0.999\linewidth]{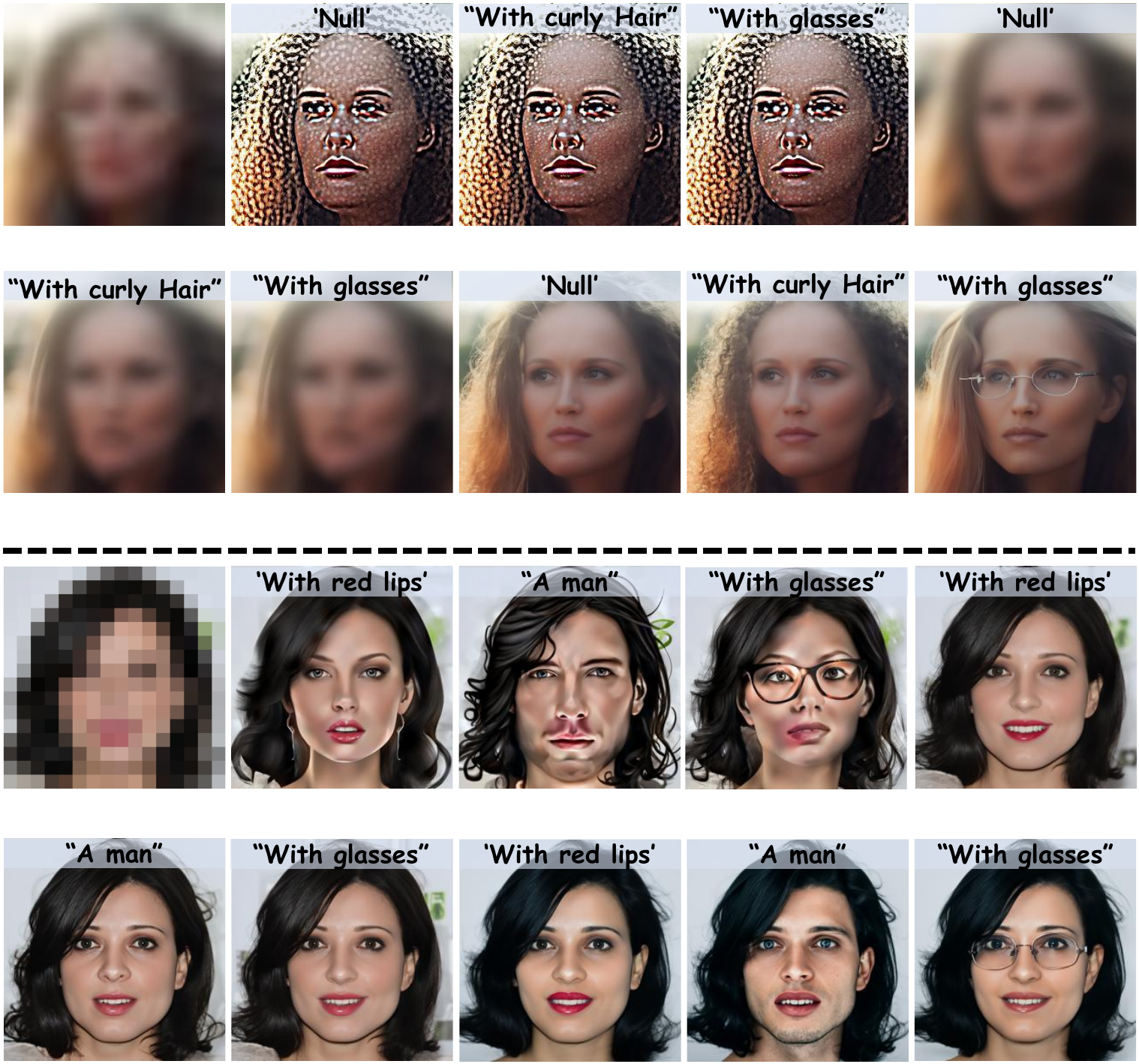}
\put(5.2,71.2){\color{black}{\fontsize{6.5pt}{1pt}\selectfont Blind LR}}
\put(20.7,71.2){\color{black}{\fontsize{6.5pt}{1pt}\selectfont T2I+DDNM\cite{wang2022zero}}}
\put(40.7,71.2){\color{black}{\fontsize{6.5pt}{1pt}\selectfont T2I+DDNM\cite{wang2022zero}}}
\put(60.7,71.2){\color{black}{\fontsize{6.5pt}{1pt}\selectfont T2I+DDNM\cite{wang2022zero}}}
\put(80.8,71.2){\color{black}{\fontsize{6.5pt}{1pt}\selectfont T2I+IIGDM\cite{wang2022zero}}}

\put(0.8,47.3){\color{black}{\fontsize{6.5pt}{1pt}\selectfont T2I+IIGDM\cite{wang2022zero}}}
\put(20.8,47.3){\color{black}{\fontsize{6.5pt}{1pt}\selectfont T2I+IIGDM\cite{wang2022zero}}}
\put(46.8,47.3){\color{black}{\fontsize{6.5pt}{1pt}\selectfont \textbf{Ours}}}
\put(68.8,47.3){\color{black}{\fontsize{6.5pt}{1pt}\selectfont \textbf{Ours}}}
\put(87.3,47.3){\color{black}{\fontsize{6.5pt}{1pt}\selectfont \textbf{Ours}}}

\put(3.5,21.4){\color{black}{\fontsize{6.5pt}{1pt}\selectfont Bicubic LR}}
\put(20.7,21.4){\color{black}{\fontsize{6.5pt}{1pt}\selectfont T2I+DDNM\cite{wang2022zero}}}
\put(40.7,21.4){\color{black}{\fontsize{6.5pt}{1pt}\selectfont T2I+DDNM\cite{wang2022zero}}}
\put(60.7,21.4){\color{black}{\fontsize{6.5pt}{1pt}\selectfont T2I+DDNM\cite{wang2022zero}}}
\put(80.8,21.4){\color{black}{\fontsize{6.5pt}{1pt}\selectfont T2I+IIGDM\cite{wang2022zero}}}

\put(0.8,-2.5){\color{black}{\fontsize{6.5pt}{1pt}\selectfont T2I+IIGDM\cite{wang2022zero}}}
\put(20.8,-2.5){\color{black}{\fontsize{6.5pt}{1pt}\selectfont T2I+IIGDM\cite{wang2022zero}}}
\put(46.8,-2.5){\color{black}{\fontsize{6.5pt}{1pt}\selectfont \textbf{Ours}}}
\put(68.8,-2.5){\color{black}{\fontsize{6.5pt}{1pt}\selectfont \textbf{Ours}}}
\put(87.3,-2.5){\color{black}{\fontsize{6.5pt}{1pt}\selectfont \textbf{Ours}}}

\end{overpic}
\vspace{-1mm}
   \caption{(\textbf{Top}): Although sampling strategies like DDNM~\cite{wang2022zero} and IIGDM~\cite{song2023pseudoinverse} have the potential to be combined with T2I model~\cite{saharia2022photorealistic} for text-aligned reconstruction, their dependence on known degradation limits performance under blind settings. (\textbf{Bottom}): Even with known degradation, existing strategies~\cite{wang2022zero,song2023pseudoinverse} combined with T2I~\cite{saharia2022photorealistic} still struggle to generate high-quality faces that align well with textual prompts. In contrast, our MCS performs well.}
\label{fig: compare_sample}
\vspace{-1mm}
\end{figure}

\vspace{-4mm}
\paragraph{Effectiveness of Reverse Measurement.}
As shown in Table~\ref{tab: reply}, removing our reverse measurement leads to a noticeable drop in response rate. Additionally, compared to methods like DPS and IIGDM, which rely on posterior gradient corrections even under known degradations, their results may suffer from semantic drift due to noise interference, leading to a decrease in response rate. Although DDNM directly projects intermediate samples onto degradation manifolds, leading to high prompt response rates, this approach may cause stylized artifacts, resulting in an oil painting–like appearance, as shown in Fig.~\ref{fig: compare_sample}. In contrast, our method corrects textual prompts in advance, offering clear guidance and enabling results of more diverse and text-aligned high-quality faces. Moreover, we observe that some guided sampling-based methods, such as DDNM~\cite{wang2022zero} and DPS~\cite{chung2022diffusion}, when combined with T2I models, can generate text-aligned reconstructions. As illustrated in Fig.~\ref{fig: compare_sample}, we show their results on a degraded face image. However, these methods rely on accurate knowledge of degradation processes, thus producing low-quality outputs when such priors are unavailable. In contrast, our method achieves high-quality, text-aligned face reconstruction without requiring explicit degradations.

\begin{figure}[t]
\begin{overpic}[width=1\linewidth]{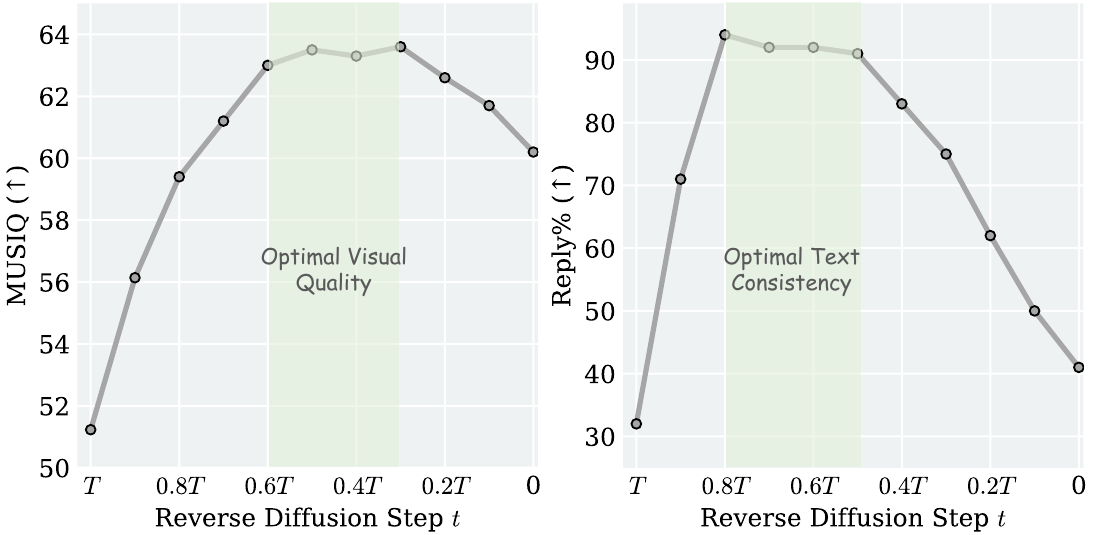}
\end{overpic}
\vspace{-3mm}
   \caption{The boundary between forward and reverse observation at $[0.6T, 0.5T]$ achieves the optimal balance between visual quality and diversity on $\times 8$ CelebA-HQ, demonstrating the effectiveness of the selection mechanism designed based on Fig.~\ref{fig: select_reason}.}
\label{fig: select_aba}
\vspace{-1mm}
\end{figure}

\vspace{-4mm}
\paragraph{Balance of Forward and Reverse Measurement.} 
Our method incorporates both forward and inverse measurements, which inherently pursue conflicting goals. The forward term (\(\mathcal{L}_1\)) encourages alignment with GT, while the reverse term (\(\mathcal{L}_2\)) promotes diversity by exploring broader solution spaces. As shown in Table~\ref{tab: Ratio}, a 1:1 weighting achieves the best trade-off between reconstruction quality and response rate. Heavier emphasis on \(\mathcal{L}_2\) improves diversity but may lead to unrealistic facial structures, while excessive \(\mathcal{L}_1\) suppresses diversity and limits the expressiveness of the generative prior.

\vspace{-3mm}
\paragraph{Effectiveness of Selection Mechanism.} 
In Sec.~\ref{subsec:Method}, statistical analysis shows that during steps $[T, 0.6T]$, the T2I diffusion model primarily converges on structural features, while steps $[0.6T, 0]$ focus on refining details. Accordingly, the selection mechanism applies $\mathcal{L}_1$ gradients during $[T, 0.6T]$ and $\mathcal{L}_2$ gradients during $[0.6T, 0]$. As shown in Fig.~\ref{fig: select_aba}, we evaluated our selection mechanism based on BFR visual quality and diversity. The optimal balance is achieved when the transition between $\mathcal{L}_1$ and $\mathcal{L}_2$ gradient losses occurred in $[0.6T, 0.5T]$, further confirming the effectiveness of our selection mechanism, as supported by statistical analysis.

\section{Limitation and Discussion}
First, our method may struggle with fine-grained texts (\eg, subtle expressions or rare accessories), likely due to the limited granularity of text embeddings and the dominance of structural constraints. Enhancing text-vision alignment or introducing spatially grounded attention may improve it. Second, for face restoration tasks with equal input and output resolution, our method relies on an experience-based projection radius $A$. However, each input may possess its own optimal projection radius $A$. This heuristic may be suboptimal; future work could learn a degradation-aware estimator to adaptively define the projection radius.

\section{Conclusion}
In this paper, we propose MCS for text-prompted BFR, addressing the inherent one-to-many nature of the problem under extremely LQ real faces. Our method combines the flexibility of sampling-based approaches with the control of T2I diffusion models, enabling the generation of diverse HQ faces conditioned on textual prompts. By introducing Forward and Reverse Measurements, MCS ensures both structural alignment with inputs and semantic diversity across reconstructions. Experiments show that MCS outperforms existing BFR methods in terms of both diversity and quality, achieving prompt-aligned restoration in a training-free manner. This work advances BFR, opening new possibilities for controllable and personalized face restoration.

{
    \small
    \bibliographystyle{ieeenat_fullname}
    \bibliography{main}
}


\end{document}